\documentclass{article}

    \usepackage[preprint]{neurips_2023}

\usepackage[utf8]{inputenc} 
\usepackage[T1]{fontenc}    
\usepackage{hyperref}       
\usepackage{url}            
\usepackage{booktabs}       
\usepackage{amsfonts}       
\usepackage{nicefrac}       
\usepackage{microtype}      
\usepackage{xcolor}         
\PassOptionsToPackage{numbers, compress}{natbib}

\usepackage{graphicx}
\usepackage{amsmath}
\usepackage{amssymb}
\usepackage{booktabs}
\usepackage{color}

\title{Compressible and Searchable: AI-native Multi-Modal Retrieval System with Learned Image Compression}

\author{%
  Jixiang Luo\thanks{Work done at Sensetime Research} \\
  \texttt{jixiangluo85@gmail.com} \\
}

\begin{document}

\maketitle

\begin{abstract}
  The burgeoning volume of digital content across diverse modalities necessitates efficient storage and retrieval methods. Conventional approaches struggle to cope with the escalating complexity and scale of multimedia data. In this paper, we proposed framework addresses this challenge by fusing AI-native multi-modal search capabilities with neural image compression. First we analyze the intricate relationship between compressibility and searchability, recognizing the pivotal role each plays in the efficiency of storage and retrieval systems.  Through the usage of simple adapter is to bridge the feature of Learned Image Compression(LIC) and Contrastive Language-Image Pretraining(CLIP) while retaining semantic fidelity and retrieval of multi-modal data. Experimental evaluations on Kodak datasets demonstrate the efficacy of our approach, showcasing significant enhancements in compression efficiency and search accuracy compared to existing methodologies. Our work marks a significant advancement towards scalable and efficient multi-modal search systems in the era of big data.
\end{abstract}

\section{Introduction}

With the exponential growth of digital content across various modalities, efficient storage and retrieval of multimedia data have become imperative. Traditional methods often struggle to cope with the increasing volume and complexity of this data. In recent years, there has been significant progress in leveraging neural network to address these challenges. Two key advancements that have garnered substantial attention are LIC and content-based image retrieval (CBIR). 

LIC techniques, enabled by deep neural networks, have revolutionized traditional compression methods. Unlike conventional compression algorithms~\cite{jpeg, bpg, bross2021overview},  that rely on handcrafted heuristics including discrete consine transformation(DCT), quantization and inverse DCT, LIC algorithms utilize neural networks to learn optimal representations of images~\cite{agustsson2019generative,  imagecnn7, imagecnn5, luo2020noise}. There exist three key directions for optimizing LIC. First is to enhance context modeling within the compression framework, allowing for better preservation of semantic information and contextual understanding~\cite{li2020spatial, yuan2021learned, imagecnn6, elic}, Secondly, subjective image quality is another key aspect by investigating techniques to improve the perceptual quality of compressed images while maintaining high compression ratios~\cite{mentzer2020high, he2022po, luo2024super}. Lastly, the strategies for model lightweighting~\cite{yu2022evaluating, liao2022efficient}, aiming to reduce the computational complexity and memory footprint of neural compression models without sacrificing performance.This enables them to achieve higher compression ratios while preserving essential semantic information. By exploiting the intrinsic structure of images, learned compression techniques pave the way for more efficient storage and transmission of visual data.

On the other hand, a regular pipeline for image retrieval often includes feature extraction, embedding and aggregation Traditional methods with CBIR often rely on handcrafted features and similarity metrics to retrieve images based on visual content. Recently learning based methods~\cite{lowe2004distinctive, sivic2003video, yue2015exploiting, liu2015deepindex, bai2018regularized, tan2021instance} have been emerging to prompt image CBIR. Moreover, the advent of large-scale pre-trained models has profoundly influenced retrieval tasks, even zerp-shot retrieval. These models like language pre-trained models ~\cite{devlin2018bert, radford2019gpt2} and multi-modal models ~\cite{clip, he2020moco, li2022blip, ramesh2021dalle}, trained on vast amounts of data, capture rich semantic information that can be leveraged for various retrieval tasks. With these models, individuals can perform not only text-to-image searches but also image-to-text searches and even image-to-image searches. This capability opens up new possibilities for cross-modal retrieval, allowing users to explore multimedia data in more versatile and intuitive ways.

In this paper, we propose a unified framework that harnesses the synergies between LIC and zero-shot retrieval for efficient multi-modal search. Our approach leverages the CLIP framework, which has demonstrated remarkable capabilities in learning joint representations of images and text. By integrating learned compression techniques with zero-shot retrieval capabilities, we aim to develop a system that enables efficient storage, retrieval, and cross-modal search of multimedia data. Through empirical evaluation on benchmark datasets, we demonstrate the effectiveness of our approach in achieving superior compression efficiency and search accuracy compared to existing methodologies.

In summary, our work represents a significant step towards analysis of compressibility and searchability and capitalizes on the feature representations learned by image compression models and repurposes them for the task of image retrieval.

\section{Related Work}
\subsection{LIC and CLIP}
LIC usually consists of three parts, including encoder, decoder and context. Thus we simplify the pipeline of LIC:
\begin{align}
    & \hat{x} = g_s(Q(g_a(x))) \\ \notag
    & \mathcal{B} = AE_1(Q(g_a(x)) + AE_2(Q(h_a(g_a(x)))) \\ \notag
    & AE_1 \sim h_s(Q(h_a(g_a(x))))
\end{align},
where $g_a, g_s, h_a, h_s$ are the encoder, decoder, the hyper-encoder and hyper-decoder at context model. $Q, AE_{1,2}, \mathcal{B}$ are the round operation, arithmetic coding and bitstream. Encoder $g_a$ extracts the information from input and decoder $g_s$ utilizes this to complete reconstruction.  Here we can identify the parameters mainly from encoder  $g_a$, decoder $g_s$ and context model $h_a, h_s$. We follow the structure of ELIC~\cite{elic} to conduct the following experiments. And its loss function is to balance the reconstruction quality and size of compressed image:
\begin{equation}
    \mathcal{L} = \mathcal{R} + \lambda * \mathcal{D}
\end{equation}\label{loss:stage1}
,where $R, D$ are the rate as the size of $\mathcal{B}$, distortion term $||x, \hat{}||$ and $\lambda$ is the Lagrange multipliers. Different from the work~\cite{liu2019codedretrieval}, this paper is to conduct an AI-native system while ~\cite{liu2019codedretrieval} just is a downstraming task combined with LIC. 

CLIP~\cite{clip} treats images as tokens, enabling zero-shot predictions with natural language. By unifying textual and visual inputs, CLIP facilitates both text-to-image and image-to-image searches, allowing for seamless exploration of multi-modal data. 
\begin{align}
    e_{image} = clip.image\_encoder(x) \\ \notag
    e_{text} = clip.text\_encoder(x)    
\end{align}
,where $e_{image}, e_{text}$ are the embedding of input image and text. Besides the encoder of CLIP has two parts, $image\_encoder$ and $text\_decoder$. The process of $image\_encoder$ in CLIP is indeed highly reminiscent of the encoder of LIC. However, while CLIP tends to provide limited information, aligning with human language understanding,  encoder in LIC encapsulates all intricate details of the image. 

\subsection{Motivation}
Therefore, our motivation is to unite the image encoder of CLIP with encoder of LIC. This approach reduces the computational overhead of feature extraction during search while leveraging features of encoder in LIC for compression, consequently minimizing storage requirements for the image database. Thus the final goal of this paper is to joint optimize the representation between $Q(ga(*))$ and $image\_encoder$. The total loss function will be discussed detail with the following sections. 

\begin{figure}[t]
  \centering
  \includegraphics[width=0.9\textwidth]{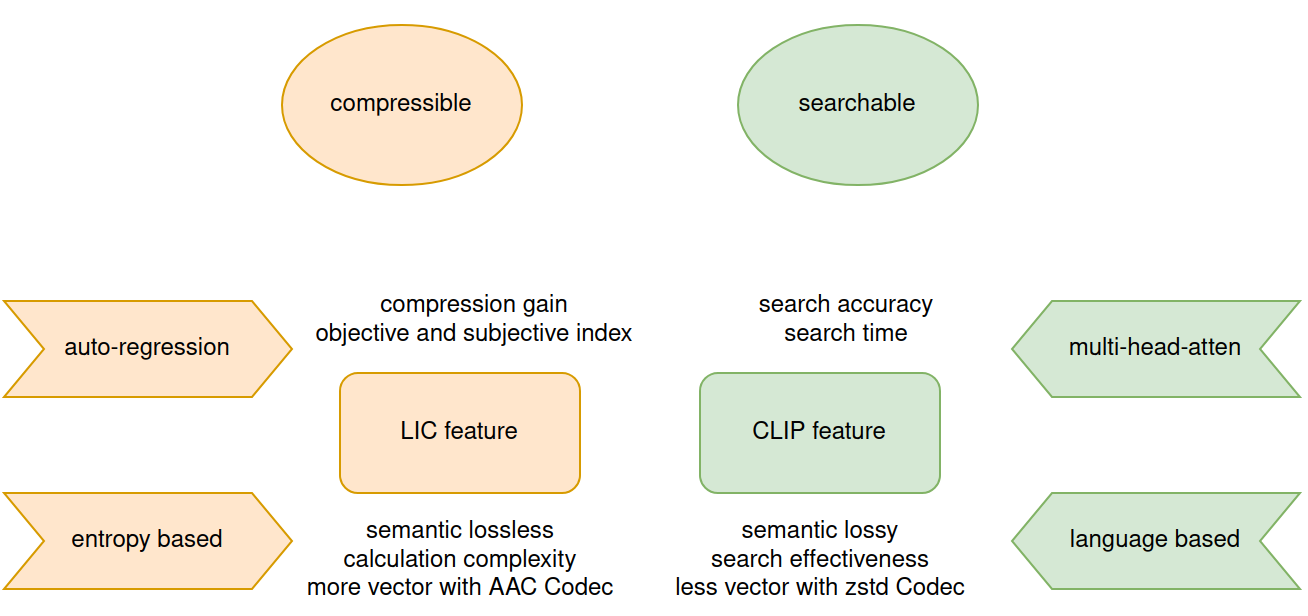}
  \caption{Compressible and Searchable}
  \label{fig:analysis}
\end{figure}

Since $image\_encoder$ of CLIP and $g_a$ can be treated as $\mathcal{M}$ to extract image feature. The capability of $image\_encoder$ has been verified to make connection image and text. However direct usage of $g_a$ for retrieval need more evidences. Thus we conduct one simple experiments to verify the effectiveness of $g_a$ to make search. 
\begin{table}[htbp]
  \centering
  \caption{The accuracy of retrieval with feature extracted by $g_a$}
  \label{verification of LIC}
  \begin{tabular}{c||c|c|c}
    \hline
    bit-rate(bpp) & 0.3 & 0.5 & 0.8\\
    hit/total & 5/24 & 7/24 & 8/24 \\
    \hline
  \end{tabular}
\end{table}
Our image dataset consists of 5000 images sampled from MSCOCO validation dataset~\cite{lin2014microsoft}, $N=5000$, the query dataset is Kodak dataset~\footnote{https://r0k.us/graphics/kodak/}, and we use the extracted features as embeddings for retrieval. The distance metric used for retrieval is cosine distance, and the standard of hitting is human score with several people. From the Tab.~\ref{verification of LIC}, it can be observed that directly using features encoded by LIC for retrieval yields very few relevant images in the top 1 results. This indicates that direct retrieval using features extracted by LIC is inefficient. Therefore, we analyze the trade-off between compressibility and searchability by analyzing its components as shown in Fig.~\ref{fig:analysis}.
The emphasis of compression and search diverges, each employing distinct structures. However, they can be analyzed using a unified approach to information processing. In compression, the context is auto-regressive, and due to its constraints on the size of compressed files, it is entropy-constrained. Its objective is to achieve higher compression gains and enhance subjective and objective quality. Conversely, the current structure utilized in search is a multi-attention mechanism aligned with human natural language understanding relative to image content comprehension. It prioritizes search accuracy and efficiency.

Furthermore, concerning compression, quality degradation may stem from the loss of details. However, ensuring the quality of reconstructed images requires that the extracted information remains semantically lossless. Maintaining such a vast amount of information to compute pairwise similarities between images, such as cosine distance, incurs considerable computational complexity. Conversely, for efficiency in search, fewer vectors are employed to represent the key semantic content within images. However, this content is lossy and subject to individual interpretation. During the storage process, LIC typically employs Adaptive Arithmetic Coding (AAC) for compression and decompression, aiming for higher compression rates. However, search systems in the database typically utilize zstd~\footnote{https://github.com/facebook/zstd} for compressing and reconstructing bitstreams, prioritizing faster processing speeds. This comes at the cost of compression performance, as the embeddings in search systems are generally short, typically ranging from 128 to 512.

\begin{figure}[htbp]
  \centering
    \includegraphics[width=0.45\textwidth]{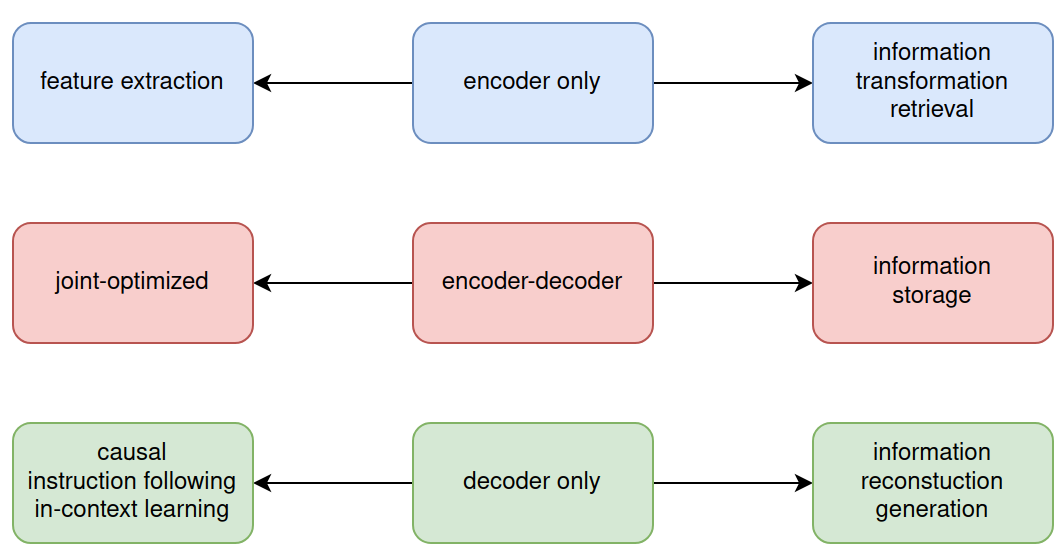}
  \includegraphics[width=0.45\textwidth]{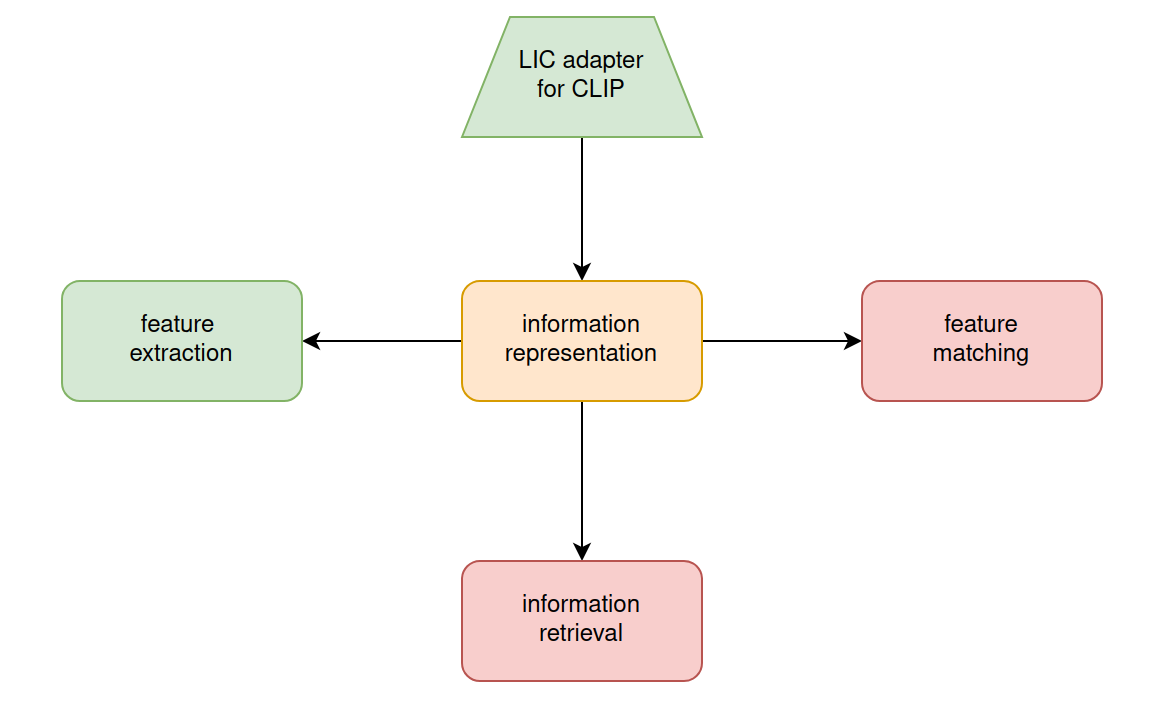}
  \caption{Left: the analysis of encoder and decoder. Right: the strategy we use to combine LIC and CLIP}
  \label{fig:adapter}
\end{figure}

Therefore, we analyze three modes of information interaction as shown in the left of Fig.~\ref{fig:adapter}: information transformation/retrieval, information storage, and information reconstruction/generation. These correspond to encoder-only, encoder-decoder, and decoder-only structures, respectively, each with their associated functions or optimization strategies. The encoder-only structure is employed for feature extraction, while the encoder-decoder structure requires joint optimization. Additionally, the prevalent decoder-only structure can primarily ensure performance through causal modeling. Leveraging techniques such as instruction following or in-context learning enables the emergence of large-scale models. Building upon above analysis, we introduce an adapter as shown in the right of Fig.~\ref{fig:adapter} to bridge the gap between LIC and CLIP features, enabling the further extraction of semantic information from LIC features to align with CLIP features.

\section{Problem Formatting}
\subsection{Image Retrieval Pipeline}
\begin{figure}[htbp]
  \centering
    \includegraphics[width=0.45\textwidth]{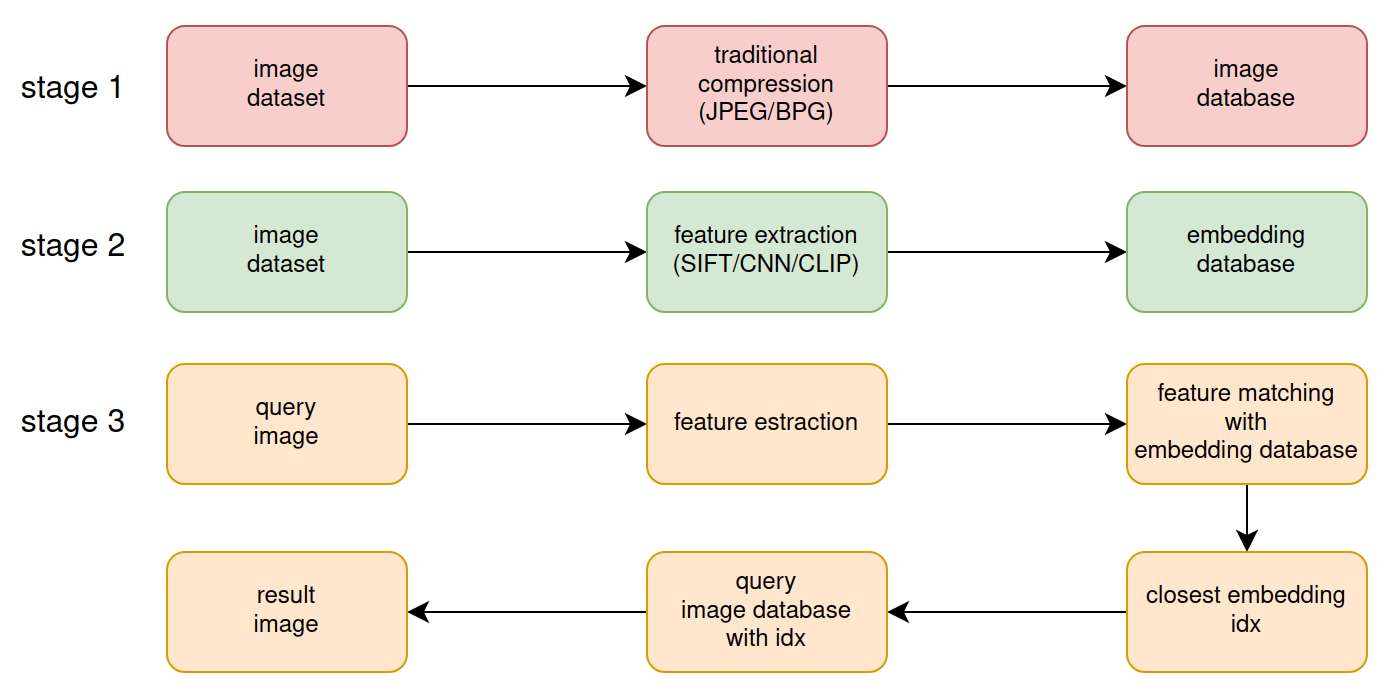}
  \includegraphics[width=0.45\textwidth]{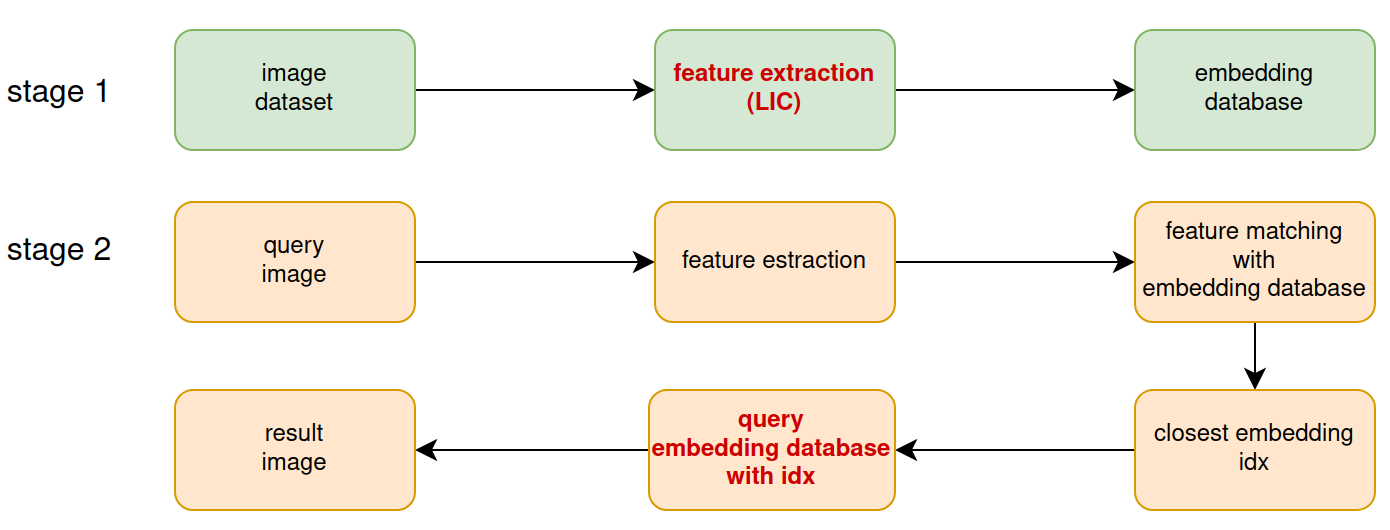}
  \caption{Left: the pipeline of traditional retrieval. Right: the pipeline of AI-native retrieval}
  \label{fig:pipeline}
\end{figure}
As show in Fig.1. The pipeline begins with an existing image database $\mathcal{I}$ with size $N$, compressed by JPEG~\cite{jpeg}, for which an embedding library $\mathcal{E} = \{M(I_i)\}, i=0,\cdots, N$ is constructed using a function $\mathcal{M}$. Upon this foundation, the search process entails converting the query image $x$ into an embedding $e$. Subsequently, the query embedding $e$ is compared against the embeddings within $\mathcal{E}$. The closest embedding $e_{target, i}$ in the library is selected, corresponding to $||e, e_{target, i}||<thr$, where $thr$ is the threshold for different metrics, and its corresponding index $i$ is determined. This index is then used to retrieve the corresponding image $I_i$ from the image database, thereby returning the result $idx$ as following: 
\begin{equation}
    idx = \arg \underset{i} {min}  ||M(x), M(I_i)||
\end{equation}

\subsection{AI-native Retrieval Pipeline}
As shown in the left of Fig.~\ref{fig:pipeline}, we streamline the process by consolidating the image database and embedding base into a unified database, housing comprehensive semantic information of the images. During reconstruction and retrieval, we only need to load embeddings extracted by LIC, then utilize an adapter for retrieval and a decoder for reconstruction. Simultaneously, the embedding extraction process for the query image also serves as a compression step. This approach accomplishes both compression and retrieval of the query image simultaneously.

\section{Methodology}
As show in Fig.~\ref{fig:adapter}, and its structure is MultiLayer Perceptron(MLP) with three convolutional layers. Thus we obtain the following relationship given image $x$:
\begin{equation}
    e = clip.image\_encode(x) \sim MLP(Q(g_a(x))
\end{equation}
However as the images input to LIC are usually multiples of 64x64, the resolution of the extracted features undergoes adaptive changes based on the image resolution after downsampling. However, since the input token length of CLIP is fixed, the adapter with single scale  cannot achieve optimal results. Therefore, we introduce three resolution alignment modules before the MLP to address this limitation. For simplicity, we define $y = Q(g_a(x)$.
\begin{equation}
    e \sim MLP(down(down(y), conv2d(down(y)), conv2d(conv2d(y))) \equiv \mathcal{M}_{LIC}(y)
\end{equation}
, where $down$ is the global pooling operation, and $conv2d$ is convolutional layer kernel with the size of $3*3$ and stide of $2$, and includes with activation function relu. And the final loss function is:
\begin{equation}
    \mathcal{L} = \mathcal{R} + \lambda *\mathcal{D} + \lambda_s * \mathcal{D}_s
\end{equation}\label{loss:stage2}
, where $\mathcal{D}_s = cosine(clip.image\_encoder(x), \mathcal{M}_{LIC}(y))$. 
Noted, it is the training loss at the second stage. 

\begin{figure}[t]
  \centering
    \includegraphics[width=0.9\textwidth]{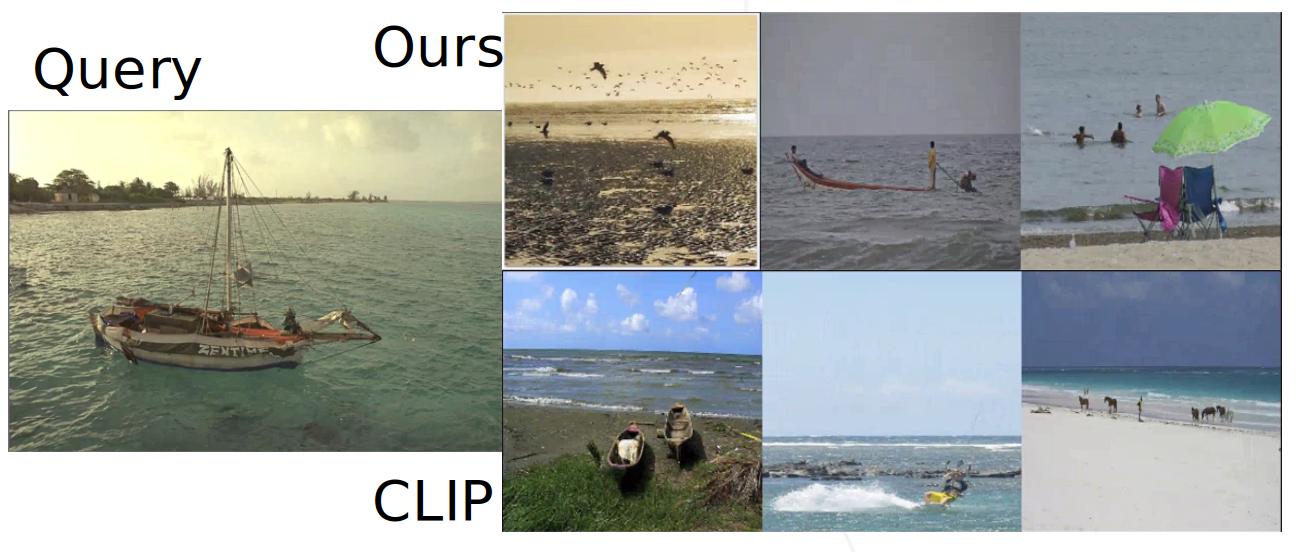}
  \caption{The retrieval result compared to CLIP}
  \label{fig:result1}
\end{figure}
\begin{figure}[t]
  \centering
    \includegraphics[width=0.9\textwidth]{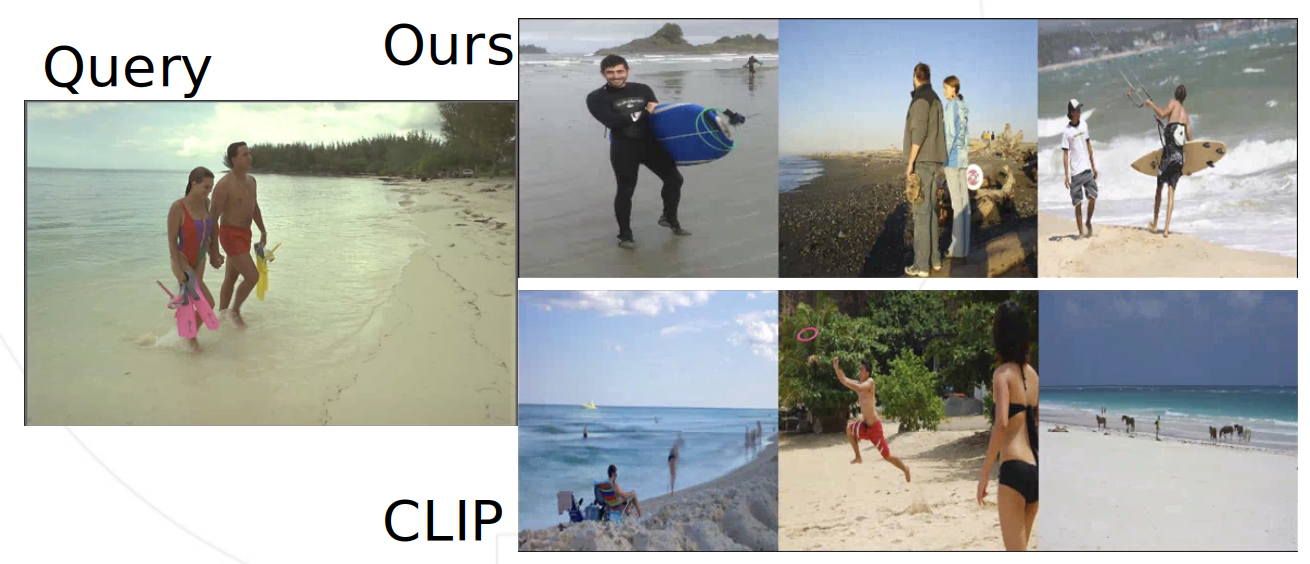}
  \caption{The retrieval result compared to CLIP}
  \label{fig:result2}
\end{figure}

\begin{table}[t]
  \centering
  \caption{The influence of different parts in LIC and the final result at last column}
  \label{ablation of LIC}
  \begin{tabular}{c||c|c|c|c|c}
    \hline
    fix &  $g_a, g_s, h_a, h_s$ & $g_a, g_s$ & $h_a, h_s$ & $None$ & $lr=le-6$ \\
    bpp & 0.5901 & 0.7655 & 1.2678 & 3.267 & 0.6002 \\
    psnr & 35.20 & 35.20 & 32.20 & 23.96 &35.02 \\
    hit/total & 6/24 & 12/24 & 12/24 & 24/24 & 24/24\\    
    \hline
  \end{tabular}
\end{table}
\section{Experiment}
The training dataset is MSCOCO, and the image database for searching is consists of 5000 images sampled from MSCOCO validation dataset. The query image is Kodak dataset.  We undergo training in two phases. In the first phase, we follow the experiments seeting as ELIC~\cite{elic} to employ loss at Eq.~\ref{loss:stage1} to train a complete LIC, comprising encoder, decoder, and context components. Subsequently, in the second phase, we integrate features extracted by CLIP for training using loss at Eq.~\ref{loss:stage2} with $\lambda_s = 0.1$. Moreover, during the second phase, we reduce the overall learning rate of LIC to 1e-6, while maintaining the learning rate of the adapter at 1e-4. Training is conducted using the Adam optimizer.

From the Fig.~\ref{fig:result1} and Fig.~\ref{fig:result2}, our system can obtain better result with regard to CLIP in top3 result with the slightly bit-rate cost from $0.5901bpp$ to $0.6002bpp$ at Tab.~\ref{ablation of LIC}. And all hit result are measured by human score with 10 people.  When an image of a boat is present in the search results at Fig.~\ref{fig:result1}, our system prioritizes showing large expanses of ocean in the top 3 results while retaining the shape of the boat. In scenarios with a prominent object in a vast, flat background, the third returned image of CLIP may lack the shape of the boat entirely, instead depicting the coastline. This discrepancy suggests a potential illusion problem with large models in CLIP, possibly stemming from differences in human understanding of image content. When searching for an image with two people at the beach at Fig.~\ref{fig:result2}, our system aims to maintain the presence of two individuals along the shore. However, CLIP lacks the capability to recognize the number of people accurately, often returning results with more than two individuals even in the top 1 result. Furthermore, despite variations in the query images, CLIP consistently returns the same result in the third image of the top 3 results, indicating limitations in the ability of CLIP to identify and understand small objects. In contrast, our system can recognize such results owing to the matching of different resolutions.

We also conduct ablation experiments to explore the impact of different components within LIC on search accuracy and compression rate, we adopt a methodology where we fix the weights of a specific component and experiment with various combinations to reflect the trade-off between compression and search.
From Tab.~\ref{ablation of LIC}, it is evident that both the encoder and decoder have a significant impact on the quality of reconstructed images and the bitrate. However, hyperscaling has a greater effect on image quality while influencing bitrate to a lesser extent. Moreover, increasing the information volume at this code point leads to an improvement in hit rate, rising from 6/24 to 12/24. Lastly, fully unleashing LIC results in an inability to ensure both bitrate and quality, further indicating that LIC features cannot be directly utilized without search.
\section{Discussion}

This paper begins by delineating the disparities between compression and search based on LIC features. Subsequently, it introduces a multi-scale adapter alongside an AI-native search system. This system extracts image embeddings by reusing encoder of LIC, reducing computational load while achieving both compression and search objectives. This provides a feasible and promising direction for the iterative update of databases. However, the paper lacks extensive experimentation to validate its efficacy further. Additionally, search experiments could benefit from specific, quantifiable metrics.

Due to the extensive data required for training large models, LIC can also be utilized for storing this data efficiently. Moreover, our system is applicable for multi-modal search and understanding, including text-to-image retrieval, image-to-image retrieval, and enhancing retrieval from large model prediction databases. Furthermore, there is significant potential to advance research in related directions, such as fine-grained understanding and retrieval of images and videos in text-to-image and text-to-video scenarios. These research avenues can contribute to the development of multi-modal databases.
{\small
\bibliographystyle{ieee_fullname}
\bibliography{egbib}
}


\end{document}